\documentclass[runningheads]{llncs}
\usepackage{booktabs}
\usepackage{multirow}
\usepackage{array}
\usepackage{amsmath}
\usepackage{cite}
\usepackage{makecell}  
\usepackage[colorlinks=true, linkcolor=blue, citecolor=green, urlcolor=magenta]{hyperref}
\usepackage{multirow} 
\usepackage{subcaption}
\usepackage{booktabs}
\usepackage{siunitx}
\usepackage{multirow}  
\usepackage{graphicx}
\usepackage{colortbl}
\usepackage{booktabs}
\usepackage{amssymb}
\usepackage{bbding}
\usepackage{pifont}
\usepackage{float}
\usepackage{lipsum}
\usepackage[T1]{fontenc}
\usepackage{pdfpages}
\usepackage{graphicx}
\usepackage{color}

\usepackage{soul}

\usepackage{xargs}
\usepackage{xstring}
\usepackage[pdftex,dvipsnames]{xcolor}  
\usepackage[colorinlistoftodos,prependcaption,textsize=tiny]{todonotes}
\newcommandx{\unsure}[2][1=]{\todo[linecolor=red,backgroundcolor=red!25,bordercolor=red,#1]{#2}}
\newcommandx{\change}[2][1=]{\todo[linecolor=blue,backgroundcolor=blue!25,bordercolor=blue,#1]{#2}}
\newcommandx{\info}[2][1=]{\todo[linecolor=OliveGreen,backgroundcolor=OliveGreen!25,bordercolor=OliveGreen,#1]{#2}}
\newcommandx{\improvement}[2][1=]{\todo[linecolor=Plum,backgroundcolor=Plum!25,bordercolor=Plum,#1]{#2}}
\newcommandx{\thiswillnotshow}[2][1=]{\todo[disable,#1]{#2}}
\begin{document}

\title{TexTAR : Textual Attribute Recognition in Multi-domain and Multi-lingual Document Images}

\titlerunning{TexTAR}

\author{Rohan Kumar\orcidID{0009-0005-8066-1240} \and
Jyothi Swaroopa Jinka\orcidID{0009-0005-1409-9034} \and
Ravi Kiran Sarvadevabhatla\orcidID{0000-0003-4134-1154}}

\authorrunning{Kumar et al.}

\institute{
Centre for Visual Information Technology,\\ International Institute of Information Technology, Hyderabad – 500032, INDIA\\
\email{\{rohan.kumar@students., jinka.swaroopa@research.,ravi.kiran\}@iiit.ac.in}}

\maketitle

\begin{abstract}
Recognizing textual attributes such as \texttt{bold}, \textit{italic}, \underline{underline}, and \st{strikeout} is essential for understanding text semantics, structure, and visual presentation. These attributes highlight key information, making them crucial for document analysis. Existing methods struggle with computational efficiency or adaptability in noisy, multilingual settings. To address this, we introduce \texttt{TexTAR}, a multi-task, context-aware Transformer for Textual Attribute Recognition (TAR). Our novel data selection pipeline enhances context awareness, and our architecture employs a 2D RoPE (Rotary Positional Embedding)-style mechanism to incorporate input context for more accurate attribute predictions. We also introduce \texttt{MMTAD}, a diverse, multilingual, multi-domain dataset annotated with text attributes across real-world documents such as legal records, notices, and textbooks. Extensive evaluations show \texttt{TexTAR} outperforms existing methods, demonstrating that contextual awareness contributes to state-of-the-art TAR performance.

\end{abstract}

\section{Introduction}

Textual Attribute Recognition (\texttt{TAR}) aims to recognize textual attributes for each word in a document image. Textual attributes like \texttt{bold}, \texttt{italic}, \texttt{underline}, \texttt{strikeout} and their combinations are essential for understanding and processing  semantic and structural elements of text in various contexts. These attributes not only emphasize meaning but also alter the strokes and visual weight of words, which play a significant role in how information is perceived. For instance, \textbf{bold text} thickens the strokes, giving words prominence and signaling importance, while \textit{italic text} introduces a slanted style, often used for emphasis or differentiation. \ul{Underlining} adds an additional stroke beneath words, drawing attention or indicating links, and \st{strikeouts} cross through words, visually signaling correction or negation. The variations in stroke thickness, orientation, and additional markings (e.g. text and background color) provide essential cues for understanding a document's hierarchy, intent, and context~\cite{agarwal-etal-2020-emplite}. They can enhance knowledge maps and information retrieval by encoding additional information within text, thus aiding tasks such as skimming and sentiment analysis~\cite{article}. They play a crucial role in user experience and readability, which is not the case for documents with plain text.

Accurately identifying textual attributes can help obtain alternative document representations, e.g. converting a document image to storage-friendly `rich-text' formats such as markdown and html. The identified attributes can also be used to enhance the capabilities of text-to-speech systems, e.g. add tonal emphasis for \texttt{bold} and communicate which portions of a document are \texttt{underlined} in document reader applications for visually challenged. 

\begin{figure*}[t]
  \centering
   \includegraphics[width=\linewidth]{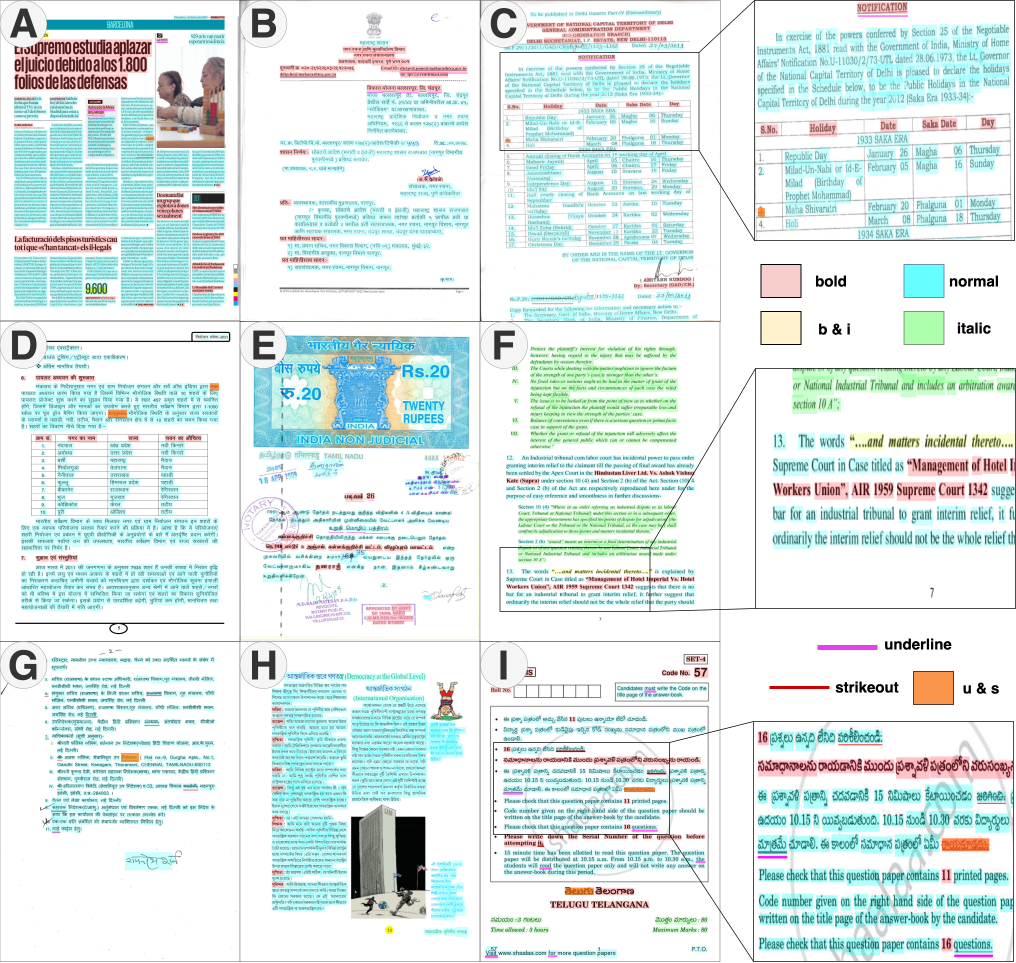}
   \caption{
   \textbf{Our MMTAD Dataset:} Examples of annotated documents from our multilingual and multidomain dataset. Note that u \& s stands for \texttt{underline \& strikeout} and b \& i stands for \texttt{bold \& italic} attribute categories.}
   \label{fig1}
\end{figure*}

\begin{figure*}[t]
  \centering
   \includegraphics[width=1\linewidth]{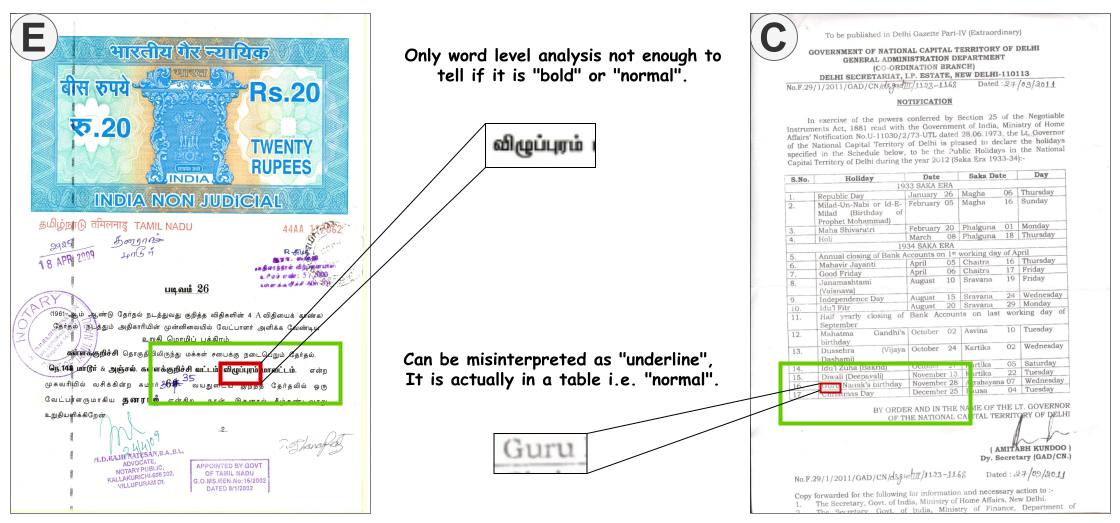}
   \caption{\textbf{Challenges in Textual Attribute Recognition (\texttt{TAR}):} (top) Challenge to distinguish \texttt{normal} and \texttt{bold} class from a single word level analysis. (bottom) Challenge to distinguish a \texttt{underline} word from a \texttt{normal} word in a table.}
   \label{fig2}
\end{figure*}

TAR is challenging due to the variety of attributes with nuanced differences, the influence of language on textual characteristics, and the intrinsic distortions in scanned documents (see Fig.~\ref{fig1}). Some of the existing approaches attempt TAR at individual word image level~\cite{mondal2021exploring,wang2015deepfont,zhong2017dropregiontraininginceptionfont}. However, they tend to underperform since textual attributes of a word cannot be determined solely by examining the word image in isolation. For instance, what appears as an underline for a word might actually be the row separator line of a table cell in a document (see Fig.~\ref{fig2}). Disambiguating such instances requires context from spatial neighborhood. In addition, many recent methods~\cite{sandu2022consent,mondal2021exploring,wang2015deepfont} tend to predict a single or a limited set of attributes, or do not explicitly consider the neighborhood context. 

To address the mentioned shortcomings, we propose TexTAR, a multi-task context-aware Transformer-based deep neural network for efficient prediction of text attributes at word level. Along with the model, we introduce a novel efficient data selection pipeline to extract the neighborhood context. To improve the network's ability in leveraging context, we introduce a novel architectural approach which utilizes 2D RoPE (Rotary Positional Embeddings)-style components within the encoder.

We also introduce a fully annotated Multilingual and Multidomain Textual Attribute Documents (\texttt{MMTAD}) dataset. Evaluation on our challenging dataset shows that TexTAR outperforms existing approaches and sets a new benchmark for text attribute recognition. The source code, pretrained models and associated material are available at
this link: \href{https://tex-tar.github.io/}{https://tex-tar.github.io/}.

\section{Related Work}

Textual Attribute Recognition(TAR) is a fine-grained recognition task for textual attributes that exploits the local features of the text strokes in the word. Researchers proposed numerous approaches for recognizing textual attributes. Previous studies \cite{inproceedings,articledevnagari,articlegurmukhi,709616} focus on the format and geometry of linguistic characters or rely on a limited set of learned font features to predict word-level text attributes such as bold and italic. Other older works \cite{inproceedingsscriptind,711217} utilize morphological operations in image processing to remove noise in the image and predict the attributes. Related classical works \cite{5715649} deal with the style transfer of the attributes from printed document text into an editable format. These approaches being rule-based focus mostly on a single language and perform adversely in the noisy multi-domain, multi-lingual document setting. 

Existing open-source libraries like Tesseract \cite{4376991} are multi-purpose text recognition libraries that recognize textual attributes like bold, italic, underline, etc. However, Tesseract predicts attributes primarily based on font style and struggles to perform in document images with subtle variations. 

In recent years, deep learning approaches \cite{sandu2022consent,nie2022taco,mondal2021exploring,wang2015deepfont} have emerged as state-of-the-art methods for textual attribute recognition tasks. DeepFont \cite{wang2015deepfont} exploits the CNN architecture for word-level feature extraction and is predominantly used for font recognition. Similarly, the approach presented in MTL \cite{mondal2021exploring} leverages multi-task learning for textual attributes such as font emphasis (bold, italic), scanning resolution, printer quality, etc. However, it does not incorporate the neighborhood context, limiting the model's ability to capture subtle feature differences effectively. The TaCo framework \cite{nie2022taco} uses contrastive learning to pretrain a customized ResNet50, integrating it with deformable DETR \cite{zhu2021deformabledetrdeformabletransformers} for both object detection and classification. While it captures neighborhood context at line level, the method is overly complex and does not generalize well. CONSENT~\cite{sandu2022consent} utilizes a transformer-based model for bold attribute recognition, leveraging neighborhood context to highlight the strengths of transformer architectures \cite{vaswani2023attention}. However, their method is narrowly focused on a single attribute and fails to address the complexities of TAR in real-world documents with multiple attributes like underline, italic, etc. In contrast, our approach introduces a data selection pipeline which captures neighbourhood context and an approach specifically designed to overcome these limitations.

\newcommand{\xmark}{\ding{55}}

\section{MMTAD Dataset}
\label{sec:dataset}
    
We introduce Multilingual and Multidomain Textual Attribute Documents \texttt{MMTAD dataset} specifically curated to recognize multiple textual attributes of a word in the document. The dataset contains $1623$ real multilingual and multidomain document images sampled from a real-world corpus with various illumination conditions, layouts, background textures, fonts, noise patterns, scanning artifacts, resolutions, color variations and other distortions. The dataset provides $1,117,716$ comprehensive word-level annotations for multiple textual attributes that cover diverse domains and languages, making it the first of its kind for multilingual and multi-domain textual attribute analysis. These documents include notices, circulars, legislative documents, land records, textbooks, and notary documents.

\begin{figure}[t]
    \centering
    \begin{minipage}{0.55\textwidth}
        \centering
        \renewcommand{\arraystretch}{1.10} 
        \setlength{\tabcolsep}{3pt}
        \begin{tabular}{|c|c|c|c|c|}
            \hline
            \multirow{2}{*}{Image} & \multicolumn{2}{c|}{$T_1$ group} & \multicolumn{2}{c|}{$T_2$ group} \\ \cline{2-5}
            & \texttt{bold} & \texttt{italic} & \texttt{underline} & \texttt{strikeout} \\ \hline
            \includegraphics[width=0.15\textwidth,height=0.020\textheight]{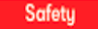} & \textcolor{red}{\xmark} & \textcolor{red}{\xmark} &\textcolor{red}{\xmark} & \textcolor{red}{\xmark}  \\ \hline
            \includegraphics[width=0.15\textwidth,height=0.020\textheight]{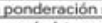} &\textcolor{red}{\xmark} & \textcolor{red}{\xmark} & \textcolor{green}{\checkmark} & \textcolor{red}{\xmark}  \\ \hline
            \includegraphics[width=0.15\textwidth,height=0.020\textheight]{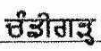} &\textcolor{green}{\checkmark} & \textcolor{red}{\xmark} & \textcolor{green}{\checkmark} & \textcolor{red}{\xmark} \\ \hline
            \includegraphics[width=0.15\textwidth,height=0.020\textheight]{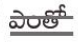} &\textcolor{red}{\xmark} & \textcolor{red}{\xmark} & \textcolor{green}{\checkmark} & \textcolor{green}{\checkmark}  \\ \hline
            \includegraphics[width=0.15\textwidth,height=0.020\textheight]{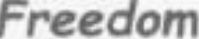} & \textcolor{green}{\checkmark} & \textcolor{green}{\checkmark} &\textcolor{red}{\xmark} & \textcolor{red}{\xmark}  \\ \hline

            \includegraphics[width=0.15\textwidth,height=0.020\textheight]{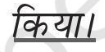} & \textcolor{red}{\xmark} & \textcolor{green}{\checkmark} &\textcolor{green}{\checkmark} & \textcolor{red}{\xmark}  \\ \hline

            \includegraphics[width=0.15\textwidth,height=0.020\textheight]{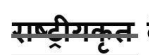} & \textcolor{green}{\checkmark} & \textcolor{red}{\xmark} &\textcolor{red}{\xmark} & \textcolor{green}{\checkmark}  \\ \hline
            
        \end{tabular}
        
    \end{minipage}
    \hfill
    \hfill
    \begin{minipage}{0.40\textwidth} 
    \centering
        \includegraphics[width=1\linewidth]{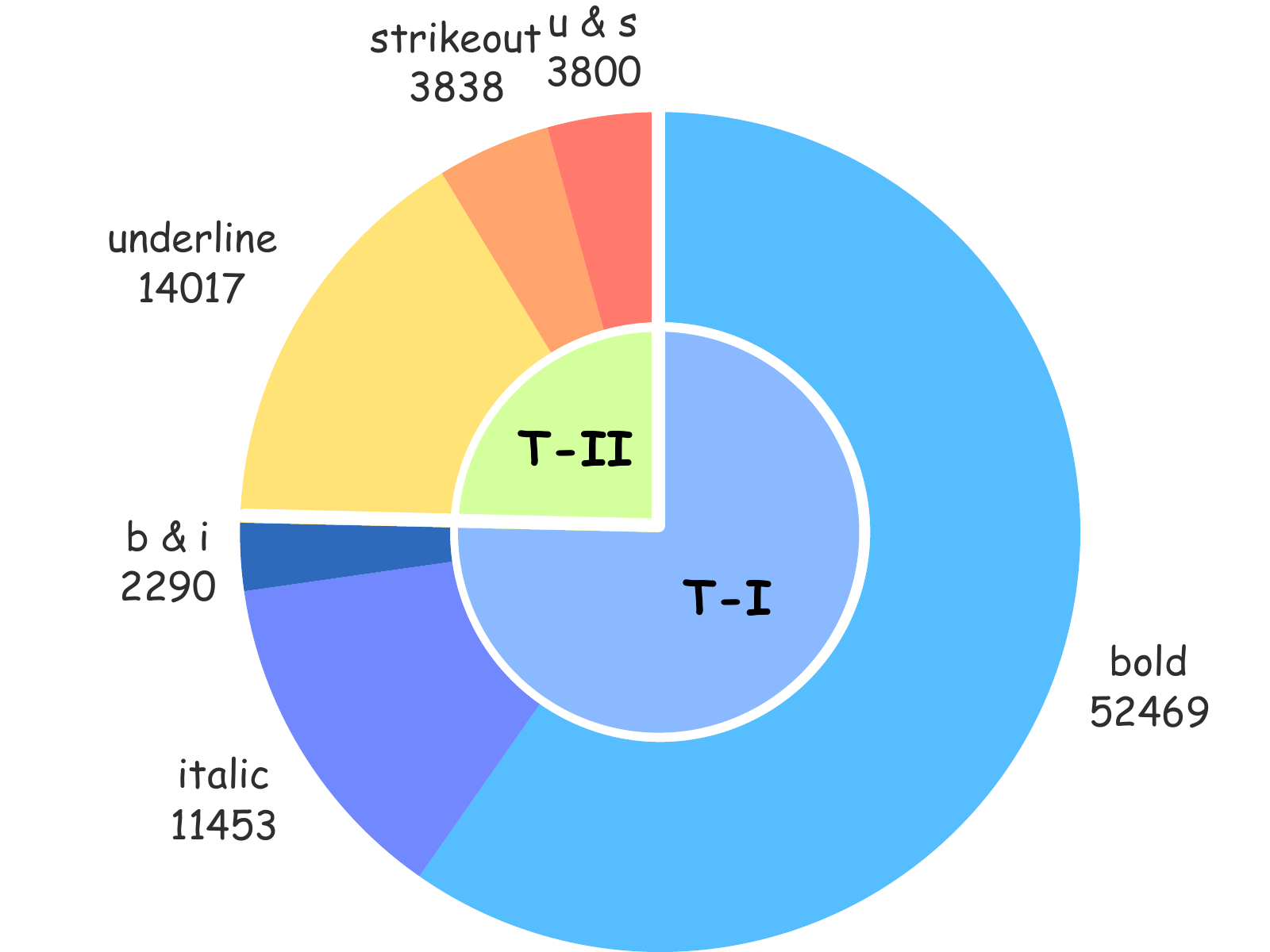} 
        \label{fig:chart}
    \end{minipage}
    \caption{\textbf{Details about textual attributes in \texttt{MMTAD} dataset:} (Left) Examples showing a subset of all possible combinations of textual attributes i.e. $T_1 \times T_2$, refer section~\ref{sec:architecture}. (Right) Chart showing the distribution of annotated attributes in our dataset.}
    \label{fig:table1}
\end{figure}

The dataset comprises documents in English, Spanish and several South Asian languages.
Each document is annotated at the word level with textual attributes categorized into two distinct groups (see Fig.~\ref{fig:table1}) . The $T_1$ group includes bold, italic and their combination. The $T_2$ group includes underline, strikeout and their combination. This grouping is motivated by the fact that attributes like bold, italic are primarily characterized by their effect on text, whereas other attributes like underline, strikeout have a prominent visual feature independent of the text. In both the groups, unadorned words which do not contain any of the group's attributes are tagged with `normal' label. Each document has an average of 300-500 annotated word bounding boxes and $87,867$ annotated (except `normal') words overall, ensuring rich and detailed metadata for textual attribute analysis. Apart from English, the dataset is inherently multilingual, with documents in Hindi (67.35\%), Telugu (8.23\%), Marathi (7.95\%), Punjabi (5.90\%), Bengali (5.35\%) and other languages (i.e. Gujarati, Tamil, Spanish) (5.22 \%) in both training and test splits. 

The number of words with specific attributes (e.g. bold) is typically much smaller than normal words. Even among words with attributes, a significant imbalance exists, e.g. bold words occur far more frequently than italic, underline and strikeout words. We augment the dataset in a context-aware manner to increase the number of samples for the italic, underline and strikeout attributes. In augmentations, we applied shear transformations to randomly italicize the words and add noisy context-aware underline and strikeout augmentations to enhance the dataset's diversity and realism. The resulting augmented dataset ensures that the samples maintain realistic appearance and variability, closely mimicking real world document distortions. 

\section{Methodology}
\label{sec:methodology}

\textbf{Overview}: Given a document image as input, our objective is to predict the corresponding textual attributes for each word - see Fig.\ref{fig:architecture}. In the first stage, we obtain word bounding boxes using a text detector\cite{doctr2021}. The output is processed through a novel data selection pipeline that generates context windows. Our method ensures that the collection of context windows for each document image includes all the words in the document and each context window contains equal number of words. In the second stage, the context windows are fed into the proposed model as input to obtain per-word text attribute predictions. 

\subsection{Stage-1: Data Selection Pipeline}
\label{sec:dsp}

\begin{figure*}[t] 
    \centering
     \includegraphics[width=\linewidth]{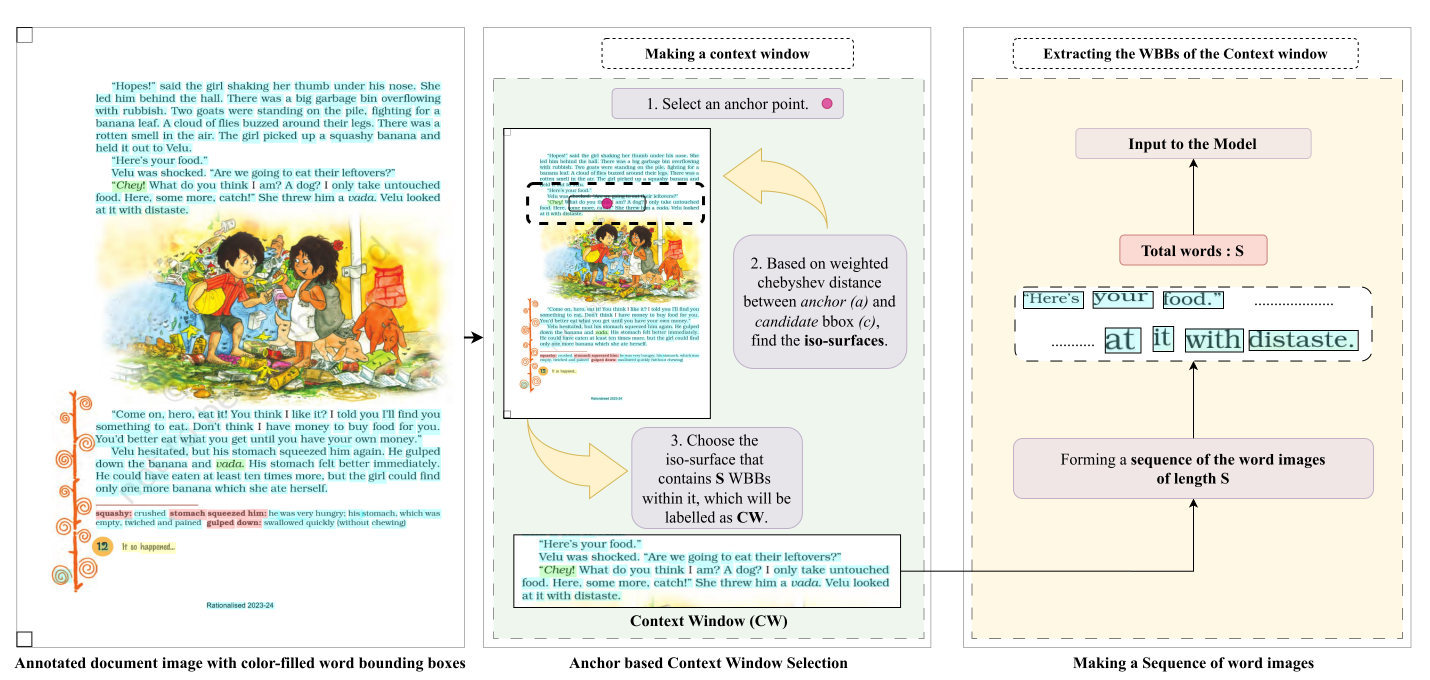}
    
    \caption{\textbf{The proposed data selection pipeline:} (a) Selection of anchor point. (b) Using the distance metric to get the nearest WBBs to the anchor point forming a \texttt{CW}. Here, the \texttt{iso-surface} is a set of WBBs equidistant from the anchor using the distance metric. Note that \texttt{CW} is a iso-surface containing $S$ words. (c) Forming a sequence input of words in \texttt{CW} for TAR model.}
    \label{fig:datapipeline} 
\end{figure*}

\begin{figure}[t] 
\centering
\includegraphics[width=\linewidth]{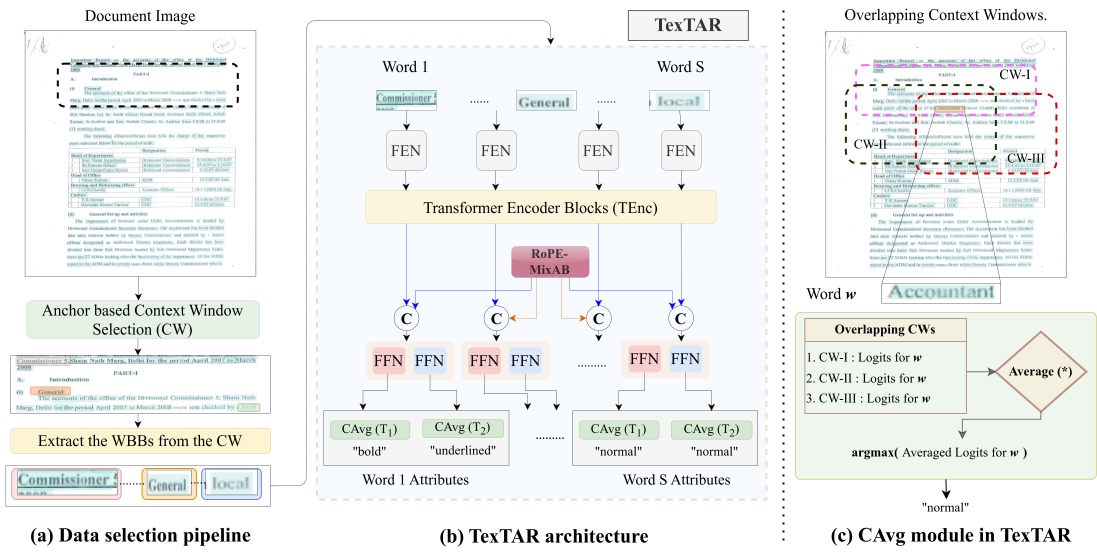}

\caption{\textbf{Overview of the method:} (a) Describes the extraction of context window $CW$ from the document. (b) Architecture of TexTAR, which takes in the sequence of word images ($CW$) and outputs the text attributes for each word in $CW$. (c) It decribes our postprocessing module i.e. \texttt{CAvg} for both $T_1$ and $T_2$.}

\label{fig:architecture} 
\end{figure}

Determining textual attributes of an isolated word solely on the basis of its appearance is a challenging task. A word that appears bold in isolation may not exhibit the same attribute when analyzed in the context of the entire document (see Fig.~\ref{fig2}-E). Similarly, a word that seems underlined individually could actually be part of a table structure (Fig.~\ref{fig2}-C). This highlights the need to incorporate the neighborhood context to accurately infer textual attributes. Most documents contain approximately 300-500 words, making it impractical to use all word bounding boxes as contextual information. Additionally, the self-attention mechanism in transformers exhibits quadratic complexity concerning input tokens, which significantly increases computational costs during both training and inference, especially for the documents with large word counts. Therefore, we propose a novel efficient data selection pipeline that extracts the relevant local context for the TAR model, allowing it to effectively capture and utilize the neighborhood's textual style information for each word (see Fig.~\ref{fig:datapipeline}).

Let  $I \in \mathbb{R}^{h \times w \times 3}$ represent a document image, where \(h\)  and \(w\)  denote the height and width of the document image respectively. A Word Bounding Box (WBB) is defined as a rectangular region enclosing a word. To incorporate contextual information, we define a Context Window (CW) around an anchor word in \(I\) . The anchor word refers to the selected word for which the CW is determined, while all other words within this window are considered as candidate words. This CW encloses the $S-1$ nearest WBBs, determined using the weighted Chebyshev distance metric, defined as: 
\[ D_{\text{Chebyshev}}(c, a) = \max (k \cdot |c_x - a_x|,m \cdot |c_y - a_y|) \] where the candidate WBB center is denoted by \(c\) and anchor WBB center is denoted by \(a\). \(k\) and \(m\) are constant weights to get a rectangular context region with $k \neq m$. 
The context window is designed to be rectangular (with width > height) to align with the document's horizontal flow of text lines. 

Each CW comprises \(S\) word images, all resized to fixed dimensions while maintaining a predetermined aspect ratio.

The number of CWs per document image vary with the number of words $S$ in the CW and the density of words in the document.

Naively generating a CW for every word would result in 300-500 context windows per document which is a substantial storage overhead. To tackle this challenge, we propose the concept of Sequential Context Windows for a given document image \( I \). This approach adopts a strategy similar to Breadth-First Search, where each Word Bounding Box (WBB) is gradually included in a CW through an iterative selection process. Initially, we mark all the WBBs in \(I\) as `unvisited'. In each iteration, an anchor word is randomly selected from the unvisited WBBs in \(I\). We construct a new CW using the \(S-1\) nearest neighbours from the anchor word. All WBBs within this CW are marked as `visited'. This ensures that they are not selected as anchor words in subsequent iterations.

This process repeats iteratively until all WBBs in the document are marked as `visited'. Our approach ensures a controlled overlap between context windows, striking a balance between storage efficiency and contextual coverage.

\subsection{Architecture}
\label{sec:architecture}

In this section, we describe the architecture of our proposed model, \texttt{TexTAR}, for Textual Attribute Recognition. We leverage the data selection pipeline (Sec.~\ref{sec:dsp}) to extract the context windows of word bounding boxes from the document image. Given a context window \( CW \) as input, the objective is to predict the textual attributes for all the words in this particular \( CW \). More formally, let the Context Window \( CW \) be defined as 
\( CW = \{ w_{1}, w_{2}, \dots, w_{S} \} \), where \( S \) denotes the sequence size of the context window. \texttt{TexTAR} maps each word in the  \( CW \) to a pair of textual attributes \( T_1 \)  and \( T_2 \)  :
\[
    \texttt{TexTAR}(w_i) \to ( T_1, T_2) \quad \forall w_i \in CW 
\]
\begin{align*}
T_1 &\in \{\texttt{normal, bold, italic, bold \& italic}\} \\
T_2 &\in \{\texttt{normal, underline, strikeout, underline \& strikeout}\}
\end{align*}

Next, we describe the component modules of \texttt{TexTAR}.

\textbf{Feature Extractor Network \texttt{(FEN)}}: 
We pass each word to a CNN network. The output from the CNN's fully connected layers provide a rich basic representation which is further refined by subsequent modules.

For each \(CW\), we obtain a set of \( S \) feature embeddings, denoted as:  
\[
CW_{emb} = \{ {FEN}(w_i) : i = 1, \dots, S \}
\]

\textbf{Transformer Encoder Blocks (\texttt{TEnc})}: The embeddings obtained from the FEN are passed to the Transformer Encoder. This encoder allows the model to learn the broader context of the window  which is crucial for disambiguating challenging cases.
Essentially, the encoder helps the model infer the correct attributes by attending to the surrounding context and learning the overall pattern of the context window (e.g. presence of a table, underline pattern that extends across all the words of a text line.(see Fig~\ref{fig1}-C)). This contextual understanding ensures that the model does not rely solely on local visual features extracted from FEN but also considers the global structure and relationships within the sequence via the Transformer attention mechanism. 
Notably, Absolute Positional Embeddings (APE)\cite{vaswani2023attention} are \textbf{not} added to \(CW_{emb}\), as they were found to reduce the performance of certain attributes (refer Table \ref{tab:ablation})
\[
T_{emb} = \{ {TEnc}({CW_{emb}^{i}})) : i = 1, \dots, S \}
\]

\textbf{Attention blocks with RoPE-Mixed (\texttt{RoPE-MixAB})}: Typically, the positional embeddings, such as sinusoidal or learnable embeddings, are added to the $QKV$ representations either at the first layer or to all the transformer layers. However, in our task, we found that this approach degrades the overall performance (Section \ref{sec:ablations}). To address this, we integrate \texttt{RoPE-Mixed} positional embeddings~\cite{heo2024rotarypositionembeddingvision} to form self-attention blocks with RoPE-Mixed PEs. We refer to these blocks as \texttt{RoPE-MixAB}. 
\texttt{RoPE-Mixed} embedding is well suited to our use case since it effectively captures diagonal frequency patterns and demonstrates strong extrapolation capabilities. This enables the model to maintain positional awareness of the WBBs surrounding a given word within a continuous 2D space (i.e. normalized 2D coordinates of WBBs in a document image) and enhances localization capabilities.
 
To balance both contextual understanding and positional encoding, we first obtain the sequence of \(T_{emb}\) from the Transformer Encoder blocks (\texttt{TEnc}) and then we apply \texttt{RoPE-MixAB} to incorporate spatial positional information into \(T_{emb}\).

\[
T_{rope} = \{ \texttt{RoPE-MixAB}(T_{emb}^{(i)}) : i = 1, \dots, S \}.
\]

After processing through the RoPE Attention Blocks, the enriched embeddings are concatenated with the original outputs of \texttt{TEnc}.  
\[
T_{final} = \{ T_{emb}^{(i)} \mathbin{\copyright}  T_{rope}^{(i)} : i = 1, \dots, S \}
\]

where $\mathbin{\copyright}$ denotes per-token concatenation.

\textbf{Dual classification heads}: The concatenated embeddings are then passed into two separate feed-forward network (FFN) heads, each designed for a specific attribute classification task group, viz. \(T_1\) and \(T_2\). Building on the discussion in Section~\ref{sec:dataset}, we employ a dual head approach to classify predictions either based on their effect on text ($T_1$) or their prominent visual characteristics independent of text ($T_2$). This strategy is empirically better than a single classification head (Section~\ref{sec:ablations}).

\textbf{CAvg module (\texttt{CAvg}):} The class logits generated by FFNs are processed within this module. For each word, the logits obtained from multiple overlapping CW predictions are averaged (see Fig.~\ref{fig:architecture}-(c)). Finally, the index corresponding to the maximum logit value determines the class label of the output attribute. This postprocessing step is carried out for both $T_1$ and $T_2$ i.e. \texttt{CAvg ($T_1$)} and \texttt{CAvg ($T_2$)} respectively.

\section{Experiments and Comparisons}
\label{sec:exps_section}

We conducted extensive experiments on our MMTAD dataset. The dataset is divided into 1005 train, 137 validation, and 481 test document images. From these images, we extract context windows to form the input data for our \texttt{TexTAR} model.

\begin{table}[t]
\centering

\setlength{\tabcolsep}{5pt}
\resizebox{0.98\textwidth}{!}{
\begin{tabular}{ll|c| c | c | c | c | c| c |c}
  \hline
  \multicolumn{2}{c|}{Methods} & \multicolumn{1}{c|}{} & \multicolumn{3}{c|}{\texttt{$T_1$ group}} & \multicolumn{3}{c|}{\texttt{$T_2$ group}} & \multicolumn{1}{c}{\textbf{Average}} \\
  &  & normal & {bold} & {italic} & {b \& i} & {underline} & {strikeout} & {u \& s} &  \\ 
  \hline
  \multirow{5}{*}{Baselines}
  & ResNet-18~\cite{he2015deepresiduallearningimage} &0.97 &0.75 &0.88 &0.77 &0.68  &0.97  &0.99 &0.86 \\
  & ResNet-50~\cite{he2015deepresiduallearningimage} &0.97 &0.74 &0.89 &0.69 &0.73  &0.98  & 0.99&0.86 \\
  &ResNeXt-101~\cite{xie2017aggregatedresidualtransformationsdeep}& 0.97 &0.77 & 0.91 &0.74 & 0.78& 0.99& 0.99 & 0.88 \\
  &EfficientNet-b4~\cite{tan2020efficientnetrethinkingmodelscaling}& 0.97 &0.75 &0.90 & 0.62& 0.75&0.98 &0.99 & 0.85\\
  \hline
  \multirow{3}{*}{Variants} & DeepFont~\cite{wang2015deepfont} & 0.97 &0.72 &0.80 &0.44 &0.64 &0.93 &0.98 & 0.78 \\
  & DropRegion\textsuperscript{\textdagger}~\cite{zhong2017dropregiontraininginceptionfont} & 0.98  & 0.77&0.90 &0.61 &0.75 &0.97 &0.99 &0.85\\
  &MTL~\cite{mondal2021exploring}& 0.97 &0.75 &0.89 &0.64 &0.70 &0.97 &0.99 &0.84\\
  &TaCo\textsuperscript{\textdagger} ~\cite{nie2022taco}& 0.97 &0.79 &0.90 &0.60 &0.78 &0.86 &0.89 &0.83\\
  &CONSENT\textsuperscript{\textdagger} ~\cite{sandu2022consent}& 0.98 &0.86 &0.93 &0.84 &0.81 &0.96 &0.98 & 0.91\\

  \hline
  \rowcolor{gray!20} 
  &\textbf{TexTAR} & \textbf{0.99} & \textbf{0.92} & \textbf{0.95} & \textbf{0.90} & \textbf{0.87} & \textbf{0.99} &\textbf{0.99} & \textbf{0.94}\\
  \hline
\end{tabular}
}

\setlength{\abovecaptionskip}{5pt}
\caption{\textbf{Comparison with state-of-the-art approaches for \texttt{TAR}:} (Section~\ref{sec:exps_section}) Note that \texttt{u \& s} is the short form for \texttt{underline \& strikeout} attribute and \texttt{b \& i} is the short form for \texttt{bold \& italic} attribute. \textsuperscript{\textdagger} Implemented by us due to unavailability of open-source code. All the above results are reported using $F_1$ metric.}
\label{tab:experiments}
\end{table}

\textbf{Implementation details :} In our experiments, each word image within a context window CW is resized to a fixed resolution of $128 \times 96$ pixels, maintaining an aspect ratio of $3:4$ . Each CW consists of a fixed number of word images ($S=125$) .

To mitigate overfitting, we incorporate dropout layers within the dual classification heads of our model and apply data augmentations, including Random Rotation, Gaussian Blur, Color Jitter, Horizontal Flip, and Random Affine, during training.

We train our model using the Adam optimizer with a learning rate of \( 0.0001 \). A weighted cross-entropy loss is employed, with empirically determined class weights of 0.25 and 0.75 for the \( T_1 \) and \( T_2 \) groups, respectively. The model is trained for 100 epochs to ensure meaningful representation learning while preventing overfitting. We empirically set the softmax temperature to 0.25. We evaluated the performance of our model using the \( F_1 \) score per attribute class.

\textbf{Training:} To ensure the consistency of outputs and the effectiveness of the proposed architecture, we adopt a two-stage training strategy. In the first stage, we train a base model comprising the Feature Extraction Network \texttt{(FEN)}, Transformer Encoder Blocks \texttt{(TEnc)}, and Dual Classification Heads. This phase allows the model to capture rich preliminary feature representations from the input word images.

In the second stage, we freeze the \textit{encoder} components from the first stage (\texttt{FEN} and \texttt{TEnc}) to preserve the learned features while mitigating positional bias. We then introduce the untrained \texttt{RoPE-MixAB} module, which is integrated by concatenating the pretrained features ($T_{emb}$) with the outputs of \texttt{RoPE-MixAB} ($T_{rope}$). Subsequently, we fine-tune the architecture while keeping the weights of \texttt{RoPE-MixAB} and the Dual Classification Heads unfrozen. This staged training approach ensures the integrity of the feature representations in the initial layers while allowing \texttt{RoPE-MixAB} to refine the embeddings without disrupting the previously learned features.

Overall, this two-stage training strategy not only stabilizes the learning process, but also ensures that the model effectively integrates positional information and attribute-specific characteristics, leading to improved performance in \texttt{TAR}.

\textbf{Comparative Analysis :} To ensure a fair evaluation, all models are trained on our \texttt{MMTAD} dataset, and their performance is assessed using the \(F_1\) score for each attribute (Refer Table~\ref{tab:ablation}). We observe that baseline feature extractors, such as ResNet-18~\cite{he2015deepresiduallearningimage}, ResNet-50~\cite{he2015deepresiduallearningimage}, and EfficientNet-B4~\cite{tan2020efficientnetrethinkingmodelscaling}, which process single-word images as input, primarily focus on attribute-specific features. As a result, their performance deteriorates significantly on attributes such as bold, underline, and their combinations, which require additional contextual information beyond single word specific feature extraction.

\begin{figure*}[!t] 
\centering
\includegraphics[width=\linewidth]{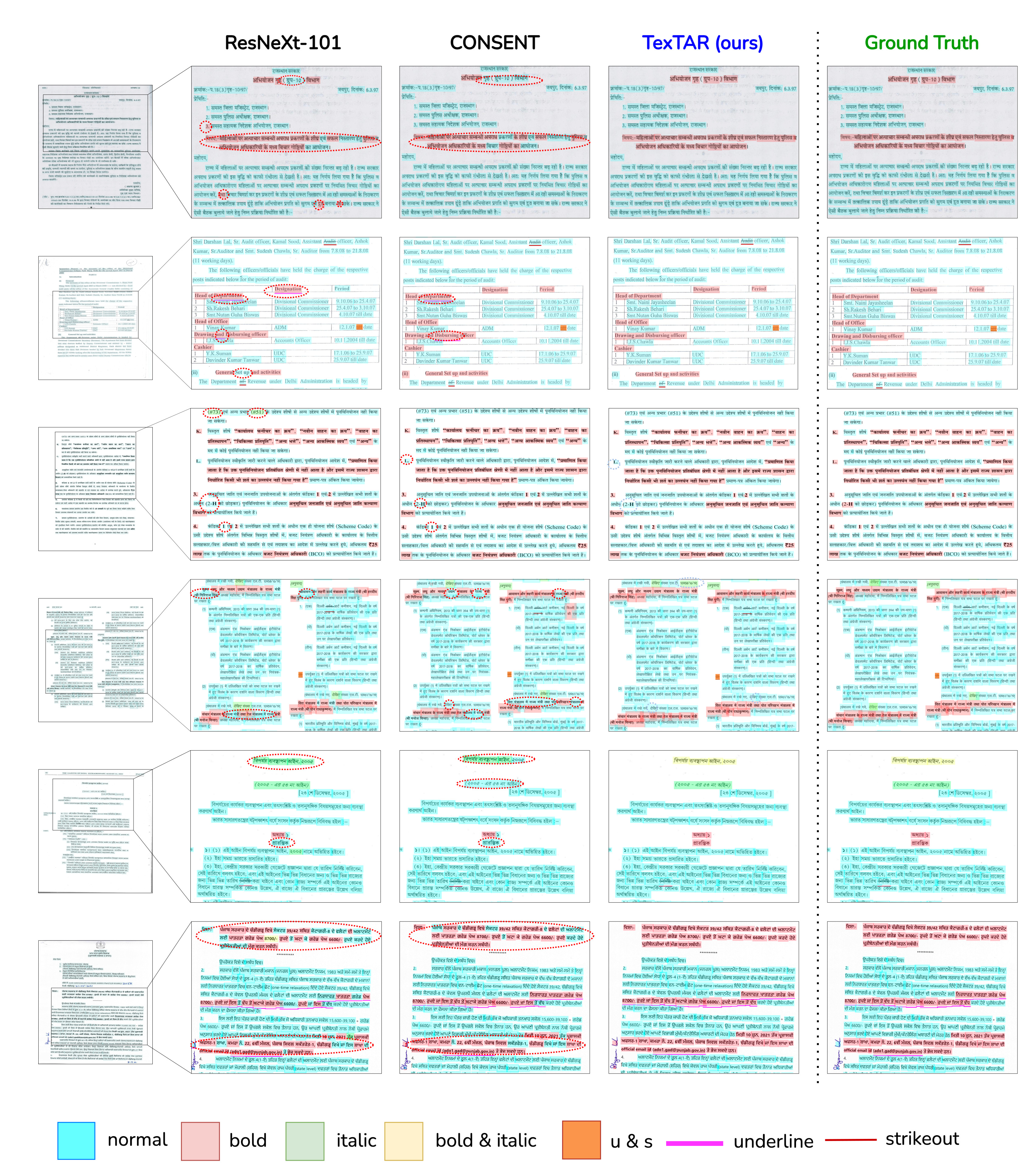}
\caption{\textbf{Qualitative Comparisions:} Visualization of results for a subset of baselines and variants in comparision with TexTAR (ours).The \textcolor{red}{red dotted circles} highlight errors made by other models, while \textcolor{darkgray}{dark gray dotted circles} indicate errors by ours, relative to the ground truth.}
\label{fig:comparisions-qualitative}
\end{figure*}

Among recent models for TAR, DeepFont~\cite{wang2015deepfont} and DropRegion~\cite{zhong2017dropregiontraininginceptionfont} exhibit subpar performance due to similar limitations. Although CONSENT~\cite{sandu2022consent} improves these baselines, it struggles to capture both positional dependencies and subtle feature differences in its proposed architecture, with a single head classifier. The TaCo~\cite{nie2022taco} model, which formulates text attribute recognition as a joint object detection and classification task, introduces dependencies on its own word detection for performance on TAR. Furthermore, its reliance on line detection as a preliminary step introduces additional bottlenecks that hinder overall performance on our \texttt{MMTAD} dataset. Qualitative comparison of TexTAR with the next best models can be seen in Fig.~\ref{fig:comparisions-qualitative}.

\subsection{Ablation Study}
\label{sec:ablations}

\begin{figure}[!t]
        \centering
        \includegraphics[width=\linewidth]{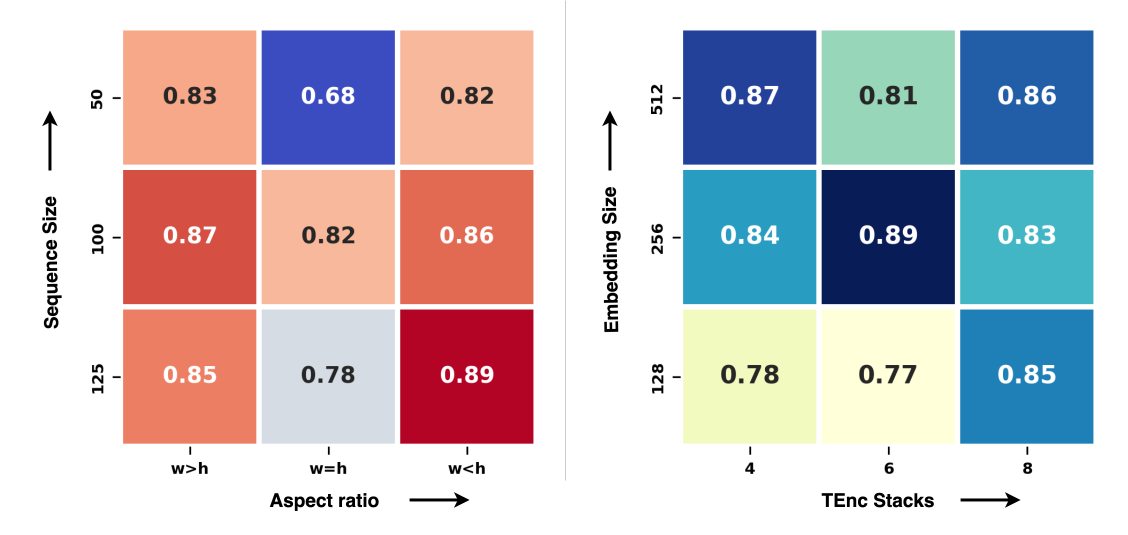}
        \caption{\textbf{Ablation experiments for \texttt{TexTAR-base-DH} model}: (left) The heatmap shows the $F_1$-scores obtained by varying Sequence size $S$ and Aspect Ratio of the word image in $CW$. (right) The heatmap shows the F1-scores obtained by varying Embedding size for word representation and the number of stacks in $\texttt{TEnc}$.}
        \label{fig:ablationTexTAR-base-DHchart}
\end{figure}

\begin{table}[t]
\centering

\setlength{\tabcolsep}{5pt}
\resizebox{0.98\textwidth}{!}{
\begin{tabular}{c| ll|c | c | c | c | c | c| c}
\hline
    & \multicolumn{2}{c|}{Ablations on TexTAR-base} & \multicolumn{1}{c|}{} & \multicolumn{3}{c|}{\texttt{$T_1$ group}} & \multicolumn{3}{c}{\texttt{$T_2$ group}}\\
  && & normal & {bold} & {italic} & {b \& i} & {underline} & {strikeout} & {u \& s} \\ 
  \hline
  
  \multirow{2}{*}{I} && TexTAR-base &0.98 &0.85 &0.93 &0.74 &0.79 &0.97 &0.99  \\
  && TexTAR-base-DH &0.99 &0.90 &0.94 &0.86 &0.87 &0.98 &0.99 \\
  \hline

  \multirow{3}{*}{II} && TexTAR-APE-DH &0.98 &0.89 &0.92 &0.72 &0.87 &0.99 &0.99 \\
  
  && TexTAR-LPE-DH &0.99 &0.90 &0.93 &0.84 &0.84 &0.98 &0.99\\
  && TexTAR-RoPE-Mix-DH &0.99 &0.90 &0.93 &0.86 &0.83 &0.99 &0.99 \\
  \hline

  \multirow{2}{*}{III} && TexTAR-base-APE-DH &0.99 &0.91 &0.95 &0.88 &0.87 &0.98 &0.99\\
   && TexTAR-base-LPE-DH &0.99 &0.91 &0.94 &0.87 &0.87 &0.98 &0.99\\
   \hline
   \multirow{2}{*}{IV} && TexTAR (E2E) &0.98 &0.82 &0.89 &0.65 & 0.83 &0.98 &0.99\\
   && TexTAR (w/o $\mathbin{\copyright}$) &0.99 &0.92 &0.94 &0.86 & \textbf{0.88} &0.99 &0.99\\
   \hline
    \rowcolor{gray!20} 
    &&\textbf{TexTAR} & \textbf{0.99} & \textbf{0.92} & \textbf{0.95} & \textbf{0.90} & 0.87 & \textbf{0.99} &\textbf{0.99}\\
  \hline
\end{tabular}}
\setlength{\abovecaptionskip}{5pt}
\caption{\textbf{Ablative studies for TexTAR:} (Section~\ref{sec:ablations}) Based on the notations in table, TexTAR is equivalent to \texttt{TexTAR-base-RoPE-Mix-DH}. Note that \texttt{u \& s} is the short form for \texttt{underline \& strikeout} attribute and \texttt{b \& i} is the short form for \texttt{bold \& italic} attribute. All the above results are reported using $F_1$ metric.}
\label{tab:ablation}
\end{table}

In this study, we assess the impact of the Dual Classification Heads, the \texttt{RoPE-MixAB} module and Concatenation procedure ($\copyright$) in our model architecture. We conduct experiments on the \texttt{MMTAD} test dataset. Our analysis focuses on an important subset of textual attributes, including \texttt{bold}, \texttt{italic}, \texttt{underline}, and \texttt{bold \& italic}. Our evaluation metric is the \(F_1\) score.

We establish a \texttt{TexTAR-base} model,  using \texttt{FEN}, \texttt{TEnc} and a single classification head having number of output nodes equal to all possible combinations of attributes ($T_1 \times T_2$). This architecture serves as the foundation for our ablations.

\textbf{Effect of Dual Classification Heads :} We analyze the impact of Dual Classification Heads in our architecture. A performance decline is observed in \texttt{TexTAR-base} (Table~\ref{tab:ablation}-I), primarily due to the skewed distribution of cross-combinations in the $T_1$ and $T_2$ groups ($T_1 \times T_2$), such as \texttt{italic} and \texttt{underline} or \texttt{bold} and \texttt{strikeout}, within the training dataset. To address this imbalance in $T_1 \times T_2$ combinations, we introduce two separate classification heads for the $T_1$ and $T_2$ groups. This design enables multi-task learning~\cite{mondal2021exploring}, improving overall performance by effectively increasing the number of samples per attribute while restricting gradient flow within each group. By preventing feature distortions in their independent heads, the Dual classification head (\texttt{DH}) approach i.e. \texttt{TexTAR-base-DH} (Table~\ref{tab:ablation}-I) achieves superior performance compared to a single-head counterpart. Fig.~\ref{fig:ablationTexTAR-base-DHchart} presents ablation studies on \texttt{TexTAR-base-DH}, showcasing its optimal performance before incorporating positional information in subsequent experiments.

\begin{figure}[!t]
        \centering
       \includegraphics[width=\linewidth]{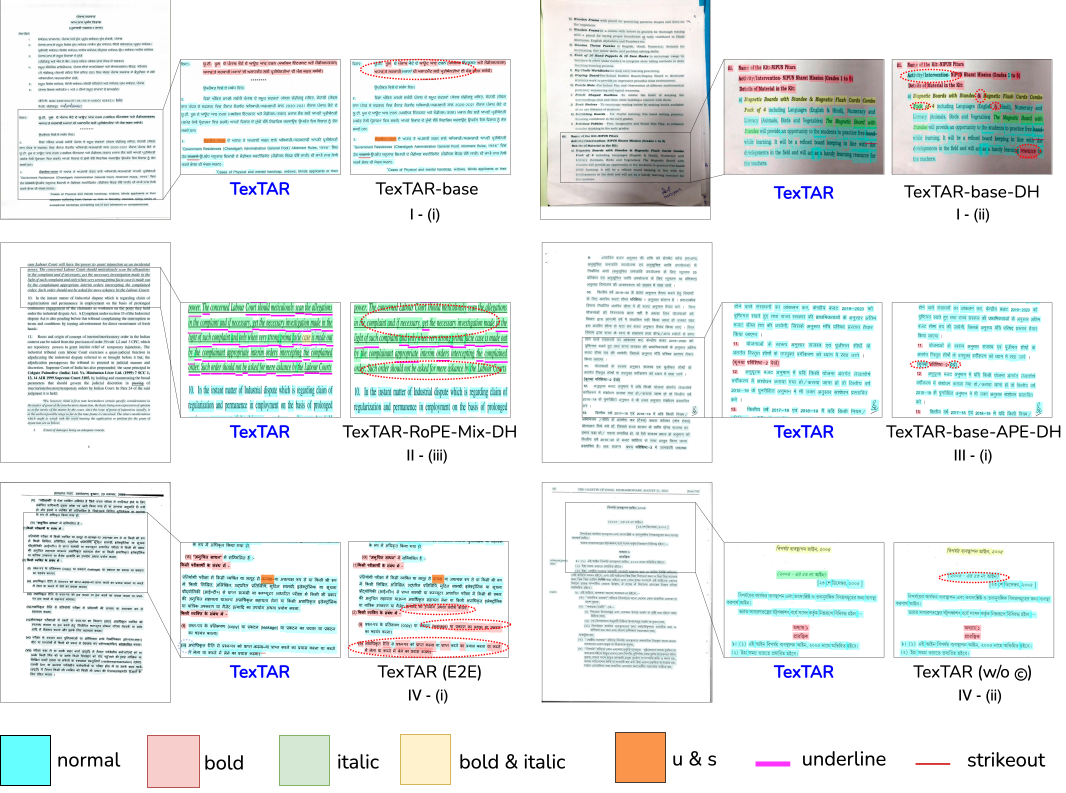}
       \caption{\textbf{Qualitative examples for ablation studies:} Examining the qualitative improvements for \texttt{TexTAR} over its ablation experiments. The \textcolor{red}{red dotted circles} highlight errors made by other ablations, while \textcolor{darkgray}{dark gray dotted circles} indicate errors by ours, relative to the ground truth. Note that the numbering \texttt{i-(j)} (below ablation name) denotes the \texttt{j}th row of \texttt{i}th section in Table~\ref{tab:ablation}.}
        \label{fig:ablationsQualitative}
\end{figure}

\textbf{Effect of RoPE-MixAB module :} We analyze the impact of positional embeddings on the predictions both qualitatively and quantitatively (see Fig~\ref{fig:ablationsQualitative}, Table~\ref{tab:ablation}). For ablative studies, we consider various positional embeddings like APE~\cite{vaswani2023attention}, Learnable PE (LPE), and RoPE-Mix~\cite{heo2024rotarypositionembeddingvision} for 2D-setting. The \texttt{RoPE-MixAB} module consists of Attention Blocks (\texttt{AB}) with RoPE-Mix PE. Here, \texttt{AB} consists of $l$ self-attention-based transformer encoder layers, where each layer can incorporate positional embeddings (PE) into the attention module. We conduct experimentations based on two training strategies : (1) End to End training of whole architecture (E2E) and (2) Pretrain, then Finetune. In the E2E strategy, our training pipeline components include \texttt{FEN}, \texttt{AB} ($l$=6) with particular PE and \texttt{DH}. We perform experiments on \texttt{AB} with different PEs like APE, LPE and RoPE-Mixed (Table~\ref{tab:ablation}-II). Though \texttt{TexTAR-RoPE-Mix-DH} dominates over others, but the \texttt{underline} performance significantly drops (0.87 to 0.83) and \texttt{italic} performance also decreases slightly (0.94 to 0.93). Further, to preserve the pre-trained features of \texttt{TexTAR-base-DH}, we adopt the strategy of fine-tuning the \texttt{AB} with PE variant keeping the training pipeline as \texttt{FEN}, \texttt{TEnc}, \texttt{AB} ($l$=2) with particular PE and \texttt{DH}. The pretrained weights for \texttt{FEN}, \texttt{TEnc} are loaded and frozen, while we fine-tune the \texttt{AB} with PE variant and \texttt{DH}. We observe that the different PE variants i.e. \texttt{TexTAR-base-APE-DH} and \texttt{TexTAR-base-LPE-DH} (Table-~\ref{tab:ablation}-III) exhibit a good performance over the E2E strategy. But, \texttt{TexTAR} (i.e. a RoPE-Mix variant) outperforms both qualitatively and quantitatively (see Fig~\ref{fig:ablationsQualitative}). In Table~\ref{tab:ablation}-IV, \texttt{TexTAR (E2E)}, trained E2E on the same architecture as TexTAR, shows performance drop. This justifies the dominance of pretraining and then fine-tuning strategy.

\textbf{Effect of Concatenation (\copyright) :} We conduct an experiment on TexTAR by removing the concatenation layer, referred to as \texttt{TexTAR w/o \copyright} (Table~\ref{tab:ablation}-IV). As observed in Table~\ref{tab:ablation}-IV, the performance of \texttt{italic} and \texttt{bold \& italic} drops, which may be due to the dominant positional bias in embeddings from \texttt{RoPE-MixAB} ($T^{(i)}_{rope}$). It leads to more dependence on positional information and neglecting learned attribute-specific features. To balance this, we concatenated the output from pre-trained \texttt{TEnc} ($T^{(i)}_{emb}$) and embeddings from \texttt{RoPE-MixAB} ($T^{(i)}_{rope}$), i.e. \texttt{TexTAR} which seem to outperform the other variants as well.

\textbf{Effect of Augmentation : } We observe that removing augmentations leads to a consistent drop in performance across $T_1$ \& $T_2$ attributes: bold drops from 0.92 to 0.87, italic from 0.95 to 0.93, and underline from 0.87 to 0.81. This highlights the importance of augmentation for robust style classification. Our method remains detector-agnostic and operates on bounding boxes produced by any text detector. To assess sensitivity to detector choice, we additionally evaluated Hi-SAM \cite{ye2025hi}.The improvements of TexTAR over other ablative studies are demonstrated through qualitative results in Fig.~\ref{fig:ablationsQualitative}.

\section{Conclusion}

TexTAR is a multi‐task, context‐aware Transformer for textual attribute recognition (TAR) that leverages self‐attention and 2D RoPE positional embeddings (RoPE-MixAB) to classify attributes such as bold, italic, underline, and strikeout at the word level. We introduce MMTAD, a fully annotated multilingual, multidomain dataset for TAR, and demonstrate that TexTAR accurately captures subtle typographic distinctions by encoding fixed-length Context Windows (CWs) around each word—extracted via an efficient detector ~\cite{doctr2021} producing per-window logits which our CAvg module then averages across overlapping CWs. Operating as a post-detection pipeline, our method avoids the irregularities of joint detection–classification, offers computational efficiency through localized CW chunking, and effectively incorporates local positional information via 2D RoPE. Extensive evaluation on MMTAD confirms TexTAR’s font- and language-agnostic robustness and state-of-the-art performance in nuanced text attribute prediction. By combining context awareness, spatial encoding, and post-window aggregation, TexTAR advances both accuracy and efficiency over existing approaches.

\section{Acknowledgements}

This work is supported by the Ministry of Electronics and Information Technology (MeitY), Government of India, as part of the Digital India Bhashini Mission, which aims to advance Indian language technology.

\begin{thebibliography}{10}
\providecommand{\url}[1]{\texttt{#1}}
\providecommand{\urlprefix}{URL }
\providecommand{\doi}[1]{https://doi.org/#1}

\bibitem{agarwal-etal-2020-emplite}
Agarwal, V., Ghosh, S., Ch, K., Challa, B., Kumari, S., Harshavardhana, Kandur~Raja, B.R.: {E}mp{L}ite: A lightweight sequence labeling model for emphasis selection of short texts. In: S, P.K.G., Mukherjee, S., Samal, R. (eds.) Proceedings of the Workshop on Joint NLP Modelling for Conversational AI @ ICON 2020. pp. 19--26. NLP Association of India (NLPAI), Patna, India (Dec 2020), \url{https://aclanthology.org/2020.icon-workshop.3/}

\bibitem{5715649}
Ajward, S., Jayasundara, N., Madushika, S., Ragel, R.: Converting printed sinhala documents to formatted editable text. In: 2010 Fifth International Conference on Information and Automation for Sustainability. pp. 138--143 (2010). \doi{10.1109/ICIAFS.2010.5715649}

\bibitem{article}
Brath, R., Banissi, E.: Font attributes enrich knowledge maps and information retrieval. International Journal on Digital Libraries  \textbf{18} (03 2017). \doi{10.1007/s00799-016-0168-4}

\bibitem{711217}
Chaudhuri, B., Garain, U.: Automatic detection of italic, bold and all-capital words in document images. In: Proceedings. Fourteenth International Conference on Pattern Recognition (Cat. No.98EX170). vol.~1, pp. 610--612 vol.1 (1998). \doi{10.1109/ICPR.1998.711217}

\bibitem{he2015deepresiduallearningimage}
He, K., Zhang, X., Ren, S., Sun, J.: Deep residual learning for image recognition (2015), \url{https://arxiv.org/abs/1512.03385}

\bibitem{heo2024rotarypositionembeddingvision}
Heo, B., Park, S., Han, D., Yun, S.: Rotary position embedding for vision transformer (2024), \url{https://arxiv.org/abs/2403.13298}

\bibitem{articledevnagari}
KantYadav, R., Mazumdar, B.: Detection of bold and italic character in devanagari script. International Journal of Computer Applications  \textbf{39},  19--22 (02 2012). \doi{10.5120/4792-7037}

\bibitem{doctr2021}
Mindee: doctr: Document text recognition. \url{https://github.com/mindee/doctr} (2021)

\bibitem{mondal2021exploring}
Mondal, T., Das, A., Ming, Z.: Exploring multi-tasking learning in document attribute classification (2021)

\bibitem{nie2022taco}
Nie, C., Hu, Y., Qu, Y., Liu, H., Jiang, D., Ren, B.: Taco: Textual attribute recognition via contrastive learning (2022)

\bibitem{inproceedingsscriptind}
Ramakrishnan, A.: Script independent detection of bold words in multi font-size documents (12 2013). \doi{10.1109/NCVPRIPG.2013.6776180}

\bibitem{sandu2022consent}
Sandu, I.C., Voinea, D., Popa, A.I.: Consent: Context sensitive transformer for bold words classification (2022)

\bibitem{articlegurmukhi}
Singh, H.: Detection of bold and italic character in gurmukhi script. IOSR Journal of Computer Engineering  \textbf{1},  28--31 (01 2012). \doi{10.9790/0661-0162831}

\bibitem{4376991}
Smith, R.: An overview of the tesseract ocr engine. In: Ninth International Conference on Document Analysis and Recognition (ICDAR 2007). vol.~2, pp. 629--633 (2007). \doi{10.1109/ICDAR.2007.4376991}

\bibitem{tan2020efficientnetrethinkingmodelscaling}
Tan, M., Le, Q.V.: Efficientnet: Rethinking model scaling for convolutional neural networks (2020), \url{https://arxiv.org/abs/1905.11946}

\bibitem{vaswani2023attention}
Vaswani, A., Shazeer, N., Parmar, N., Uszkoreit, J., Jones, L., Gomez, A.N., Kaiser, L., Polosukhin, I.: Attention is all you need (2023)

\bibitem{wang2015deepfont}
Wang, Z., Yang, J., Jin, H., Shechtman, E., Agarwala, A., Brandt, J., Huang, T.S.: Deepfont: Identify your font from an image (2015)

\bibitem{xie2017aggregatedresidualtransformationsdeep}
Xie, S., Girshick, R., Dollár, P., Tu, Z., He, K.: Aggregated residual transformations for deep neural networks (2017), \url{https://arxiv.org/abs/1611.05431}

\bibitem{ye2025hi}
Ye, M., Zhang, J., Liu, J., Liu, C., Yin, B., Liu, C., Du, B., Tao, D.: Hi-sam: Marrying segment anything model for hierarchical text segmentation. IEEE Transactions on Pattern Analysis and Machine Intelligence  \textbf{47}(03),  1431--1447 (2025)

\bibitem{inproceedings}
Zhang, L., Lu, Y., Tan, C.L.: Italic font recognition using stroke pattern analysis on wavelet decomposed word images. vol.~4, pp. 835--838 (01 2004). \doi{10.1109/ICPR.2004.1333902}

\bibitem{zhong2017dropregiontraininginceptionfont}
Zhong, S.H.Z., Jin, L., Zhang, S., Wang, H.: Dropregion training of inception font network for high-performance chinese font recognition (2017), \url{https://arxiv.org/abs/1703.05870}

\bibitem{zhu2021deformabledetrdeformabletransformers}
Zhu, X., Su, W., Lu, L., Li, B., Wang, X., Dai, J.: Deformable detr: Deformable transformers for end-to-end object detection (2021), \url{https://arxiv.org/abs/2010.04159}

\bibitem{709616}
Zramdini, A., Ingold, R.: Optical font recognition using typographical features. IEEE Transactions on Pattern Analysis and Machine Intelligence  \textbf{20}(8),  877--882 (1998). \doi{10.1109/34.709616}

\end{thebibliography}

\end{document}